\documentclass[lettersize,journal]{IEEEtran}
\usepackage{amsmath,amsfonts}
\usepackage{algorithmic}
\usepackage{algorithm}
\usepackage{array}
\usepackage[caption=false,font=normalsize,labelfont=sf,textfont=sf]{subfig}
\usepackage{textcomp}
\usepackage{stfloats}
\usepackage{url}
\usepackage{verbatim}
\usepackage{graphicx}
\usepackage{cite}
\usepackage{multirow}
\begin{document}

\title{A Novel Pareto-optimal Ranking Method for Comparing Multi-objective Optimization Algorithms}

\author{Amin Ibrahim, Azam Asilian Bidgoli, Shahryar Rahnamayan, and Kalyanmoy Deb
\thanks{Author1, is with the Faculty of Business and IT, Ontario Tech University, 2000 Simcoe Street North, Oshawa, Ontario, L1G 0C5, Canada, Corresponding author's e-mail: amin.ibrahim@ontariotechu.ca}
\thanks{Author2 is with Computer Science, Wilfrid Laurier University, 75 University Avenue, Waterloo, Ontario, N2L 3C5,  Canada }
\thanks{Author3 is with the Department of Engineering, Brock University, 1812 Sir Isaac Brock Way
St. Catharines, Ontario, L2S 3A1, Canada}
\thanks{Author4 is with the Department of Electrical and Computer Engineering, Michigan State
University, East Lansing, MI 48824, USA}
}
\markboth{Journal of ...}%
{Shell \MakeLowercase{\textit{et al.}}: Bare Demo of IEEEtran.cls for IEEE Journals}

\maketitle

\begin{abstract}

As the interest in multi- and many-objective optimization algorithms grows, the performance comparison of these algorithms becomes increasingly important. A large number of performance indicators for multi-objective optimization algorithms have been introduced, each of which evaluates these algorithms based on a certain aspect. Therefore, assessing the quality of multi-objective results using multiple indicators is essential to guarantee that the evaluation considers all quality perspectives. This paper proposes a novel multi-metric comparison method to rank the performance of multi-/ many-objective optimization algorithms based on a set of performance indicators. We utilize the Pareto optimality concept (i.e., non-dominated sorting algorithm) to create the rank levels of algorithms by simultaneously considering multiple performance indicators as criteria/objectives. As a result, four different techniques are proposed to rank algorithms based on their contribution at each Pareto level. This method allows researchers to utilize a set of existing/newly developed performance metrics to adequately assess/rank multi-/many-objective algorithms. The proposed methods are scalable and can accommodate in its comprehensive scheme any newly introduced metric. The method was applied to rank 10 competing algorithms in the 2018 CEC competition solving 15 many-objective test problems. The Pareto-optimal ranking was conducted based on 10 well-known multi-objective performance indicators and the results were compared to the final ranks reported by the competition, which were based on the inverted generational distance (IGD) and hypervolume indicator (HV) measures. The techniques suggested in this paper have broad applications in science and engineering, particularly in areas where multiple metrics are used for comparisons. Examples include machine learning and data mining.

\end{abstract}


\IEEEpeerreviewmaketitle

\section{Introduction}

Since many real-world problems are modeled as multi-objective optimization problems, a large number of multi-objective algorithms have been developed to tackle them~\cite{hua2021survey, asilian2022machine,sharma2022comprehensive}. In practical approaches, multi-objective algorithms are required to deal with a set of conflicting objectives, and thus finding the optimal solutions may not be easily observable\cite{du2022multi}. Accordingly, the nature of conflicting objectives generates a set of trade-off solutions as the \emph{Pareto front} solutions of a multi-objective optimization problem. On the other hand, the evaluation of the solution sets produced by the multi-objective algorithms, which presents various trade-offs among the objectives must be quantitatively appraised using a variety of measurement metrics~\cite{okabe2003critical}. While the quality evaluation of the single-objective solution is trivial (i.e., for minimization, the smaller the value of the objective, the better), measuring the quality of Pareto fronts resulting from multi-objective algorithms is complex. A Pareto front should be evaluated using aspects such as diversity, distribution, and closeness to the true Pareto front~\cite{zitzler2000comparison}. In order to assess the developed algorithms and analyze the results, a vast number of performance metrics have been proposed~\cite{7360024}. Miqing Li \emph{et al.} ~\cite{li2019quality} presented a study of 100 multi-objective optimization metrics that have been used in the specialized literature. Their paper discusses the usage, trends, benefits, and drawbacks of the most popular metrics to provide researchers with essential knowledge when selecting performance metrics.

Audet \emph{et al.}~\cite{audet2021performance} classified multi-objective performance indicators into four main categories including cardinality indicators, convergence indicators, distribution and spread indicators, and convergence and distribution indicators. Cardinality indicators such as two-set coverage~\cite{zitzler1998multiobjective} evaluate the quality of the Pareto front based on the number of non-dominated points generated by the corresponding algorithm. Convergence indicators quantify the closeness of the resulting Pareto front to the true Pareto front. Generational distance (GD)~\cite{van1999multiobjective} and inverted GD (IGD)~\cite{coello2005solving} are two well-known metrics that belong to this category. 
Distribution and spread indicators measure the distribution and the extent of spacing among non-dominated solutions. Spacing~\cite{schott1995fault} is an instance of distribution and spread indicators. Finally, the convergence and distribution indicators capture both the properties of convergence and distribution. One of the popular metrics in this category is the hypervolume (HV) indicator~\cite{zitzler1999multiobjective}.

In addition to the categorization based on the factors that each metric captures, the performance indicators can be divided into unary and binary~\cite{riquelme2015performance}. Unary metrics provide a real value after taking into account one or more of the aforementioned factors, whereas binary metrics focus primarily on the relationship between two approximation sets resulting from two algorithms to determine which one is better.

In general, there is no universally supreme metric since each performance indicator may have some strengths and limitations~\cite{nuh2021performance}. Moreover, we require at least as many indicators as the number of objectives in order to determine whether an approximate solution is \emph{better} than another (i.e., one objective vector dominates or weakly dominates another) ~\cite{zitzler2003performance}. Each developed metric captures one or sometimes several aforementioned factors at the same time and assigns a score to the result of each algorithm. This indicates that - no ideal individual metric can assess all characteristics of a Pareto front and, consequently,  considering several metrics simultaneously is crucial.  In such cases, the scores of algorithms based on each metric are computed and the overall final rank is determined. However, this can also be more complicated if the metrics conflict with each other, thus indicating the overall ranking can be more challenging. This occurs because each performance indicator may yield distinct rankings for competing algorithms. For instance,~\cite{jiang2014consistencies} investigated the contradictions of the IGD and HV indicator values when evaluating concave Pareto fronts. In literature, there exist a few publications that have studied the application of multiple performance metrics~\cite{yen2013performance, zitzler2009set}. Yen \emph{et al.}~\cite{yen2013performance}  introduced an ensemble method to rank algorithms by combining several evaluation metrics using double-elimination tournament selection. As an alternative approach, there have been some techniques to combine several metrics to present one ranking result based on a comprised indicator~\cite{yan2007diversity}. However, finding a combination technique to avoid the negative impact on the individual metrics can be another challenging issue. Furthermore, even with a combination technique, only a limited number (two or three) of indicators can be combined. From these investigations,  it is evident that multi-metric evaluation of algorithms has more benefits than any stand-alone single performance metric and the set of quality indicators has to be large enough to satisfy the reliability of the comparison~\cite{ravber2017impact}.

To address the aforementioned concerns and to ensure a fair assessment, this paper proposes a multi-metric method to rank the multi-objective algorithms based on a set of performance indicators. The proposed approach utilizes the Pareto optimality concept to tackle the issue of possible conflicts among the measurements. Each performance metric can be observed as an objective in the objective space and, consequently, algorithms are ranked based on their scores achieved from the individual metrics.  Since the Pareto optimality leads to different Pareto levels, four techniques are proposed to rank algorithms based on their contribution at each Pareto level. This technique provides a reliable ranking at the end of the process, regardless of the objectives, metrics, and algorithms employed. However, it is crucial to choose appropriate performance metrics that align with our ultimate goal. This is because the metrics we opt for play a vital role in accurately ranking competing algorithms. Furthermore, any newly developed metrics can be included as part of the assessment. A great benefit of this method is that it is parameter-free and algorithms can be evaluated on various factors, with the overall ranking generated at the end of its process.

The remaining sections of this paper are organized as follows: A background review of multi-objective optimization is provided in Section II. Section III reviews some well-known multi-objective performance indicators. Section IV presents a detailed description of the proposed multi-metric method. Section V investigates the performance of the proposed method using the 2018 CEC  competition's test problems and algorithms. Finally, the paper is concluded in Section VI.

\section{Background Review on Multi- and Many-objective optimization}
Multi-objective optimization targets handling two or three conflicting objectives and many-objective algorithms aim to tackle more than three conflicting objectives. In recent years, multi-objective optimization algorithms have been greatly expanded to tackle many-objective problems. The use of evolutionary algorithms has been very promising for solving such problems. As a population-based approach, it enables the generation of a set of solutions at each run, with each solution potentially interacting with the others to create even better solutions.  

\textbf{Definition 1. Multi-objective Optimization~\cite{asilian2022machine}}
\begin{eqnarray}
\begin{aligned}
& Min/Max\  F(\pmb x)=[f_{1}(\pmb x),f_{2}(\pmb x),...,f_{M}(\pmb x)] \\ 
&s.t. \quad L_{i}\leq x_{i}\leq U_{i}, i=1,2,...,d
 \end{aligned}
\end{eqnarray}
subject to the following equality and/or inequality constraints.
\begin{eqnarray}
\begin{aligned}
g_j(\pmb x) \leq 0 \quad j=1,2,...,J\\
h_k(\pmb x)=0 \quad k=1,2,...,K
\end{aligned}
\end{eqnarray}
where $M$ is the number of objectives, $d$ is the number of decision variables (i.e., dimension), and the value of each variable, $\pmb x_{i}$, is in interval $[L_{i},U_{i}]$ (i.e., box-constraints). $f_{i}$ represents the objective function, which should be minimized or maximized. The hard constraints that are required to be satisfied are $g_j(\pmb x) \leq 0 \quad j=1,2,...,J$ and $h_k(\pmb x)=0 \quad k=1,2,...,K$.

In multi- or many-objective optimization problems, finding an optimal solution set is far more complex than in the single-objective case. As such, a trade-off must be made between the different objectives. One way to compare the various candidate solutions is to use the concept of dominance. This involves an assessment of how one solution is better than another with regard to the objectives.

\textbf{Definition 2. Dominance Concept ~\cite{deb2002fast}}
If  $\pmb x=(x_{1},x_{2},...,x_{d})$ and  $ \acute{\pmb x}=(\acute{x}_{1},\acute{x}_{2},...,\acute{ x}_{d})$ are two vectors in a minimization problem search space, $\pmb x$ dominates $\acute{\pmb x}$ ($\pmb x\prec\acute{\pmb x}$) if and only if
\begin{eqnarray}
\begin{aligned}
&\forall i\in{\{1,2,...,M\}}, f_i(\pmb x)\leq f_i(\acute{\pmb x}) \wedge\\ 
&\exists j \in{\{1,2,...,M\}}: f_j(\pmb x)<f_j(\acute{\pmb x})
\end{aligned}
\end{eqnarray}
This concept defines the optimality of a solution in a multi-objective space. Candidate solution $\pmb x$ is better than $\acute{\pmb x}$ if it is not worse than $\acute{\pmb x}$ in any of the objectives and it has a better value in at least one of the objectives. Solutions that are not dominated by any other solution are called non-dominated solutions; they create the  Pareto front set. Multi-objective algorithms attempt to find these solutions by utilizing generating strategies/operators and selection schemes. The non-dominated sorting (NDS) algorithm~\cite{deb2002fast} is one of the popular selection strategies that work based on the dominance concept. It ranks the solutions of the population in different levels of optimality, called Pareto. 
The algorithm starts with determining all non-dominated solutions in the first rank.
 In order to identify the second rank of individuals, the non-dominated vectors are removed from the set to process the remaining candidate solutions in the same way. Non-dominated solutions of this step make the second level of individuals (second Pareto). Thereafter, the second-ranked individuals will be removed to identify the third Pareto. This process will continue until all of the individuals are grouped into different levels of Pareto.
 \section{Literature review on performance indicators}
 There are several performance metrics to assess the quality of multi and many-objective algorithms. These metrics evaluate the performance of the algorithms using aspects such as convergence, distribution, and coverage. Each metric may have its advantages and drawbacks. In this section, we review some well-known metrics that we have utilized in our experiments to design the multi-metric ranking method. Table~\ref{metricsProperties} provides an overview of the ten metrics employed in this study, detailing what they measure, the required number of parameters, and their respective advantages and disadvantages. The table primarily addresses three key aspects of a solution set: convergence (proximity to the theoretical Pareto optimal front), diversity (distribution and spread), and cardinality (number of solutions).

\textbf{Hypervolume (HV) indicator}~\cite{zitzler1999multiobjective}: HV is a very popular indicator that evaluates multi-objective algorithms in terms of the distribution of the Pareto front and the closeness to true Pareto front. HV indicator evaluates the diversity and convergence of a multi-objective algorithm. It calculates the volume of an $M$-dimensional space that is surrounded by a set of solution points $(A)$ and a reference point $r=(r_{1}, r_{2},...,r_{M})$ where $M$ is the number of objectives of the problem. Therefore, HV measures the volume of the region which is dominated by the non-dominated solutions in the objective space, relative to a reference point, $r$. A reference point is a point with worse values than a nadir point. The HV measure is defined in Eq. \ref{HV}.
\begin{equation}
\label{HV}
HV(A)=vol\ (\bigcup_{a\in A}[f_{1}(a),r_{1}]\times[f_{2}(a),r_{2}]\times...\times[f_{M}(a),r_{M}]),
\end{equation}
where $f_{i}$ represents the objective function and $a \in A$ is a point that weakly dominates all candidate solutions. Larger values of HV indicate that the Pareto front surrounds a wider space resulting in more diverse solutions and closer to the optimal Pareto front. 

\textbf{Generational Distance (GD)}~\cite{van1999multiobjective}: 
GD measures the average minimum distance between each obtained objective vector from the set $S$ and the closest objective vector on the representative Pareto front, $P$, which  is calculated as follows:
\begin{equation}
\label{GD}
GD(S,P)=\frac{\sqrt{\sum^{|S|}_{i=1}{dist(i,P)^2}}}{|S|},
\end{equation}
where $dist(i,P)$ is the Euclidean distance from an approximate solution to the nearest solution on the true Pareto front $q = 2$. A smaller GD value indicates a lower distance to the true Pareto front and consequently better performance.

\textbf{Inverted Generational Distance (IGD)}~\cite{coello2005solving}: The only difference between GD and IGD is that the average minimum distance measure is from each point in the true Pareto to those in the approximate Pareto front, so that:
\begin{equation}
IGD(S, P) = \frac{\sqrt{\sum^{|P|}_{i=1}{dist(i,S)^2}}}{|P|},
\end{equation}
where $dist(i,S)$ is the Euclidean distance between a point in the true Pareto front and the nearest solution on the approximate solution.

\textbf{Two-set Coverage (C)}~\cite{zitzler1998multiobjective}: 
This metric indicates the rate of solutions on the Pareto front of one algorithm that is dominated using solutions of another algorithm. 
\begin{equation}
C(A,B)=\frac{|\{b\in B, \text{there exists }a\in A \text{ such that }a \preceq b \}|}{|B|}
\end{equation}
For example, $C(A,B)= 0.25$ means that the approximate solutions from algorithm A dominate 25\% of the solutions resulting from algorithm B. Obviously, both $C(A,B)$ and $C(B,A)$ should be calculated for comparison.

\textbf{Coverage over the Pareto front (CPF)}~\cite{tian2019diversity}: 
This is a measure of the diversity of the solutions projected through a mapping from an $M$-dimensional objective space to an $(M-1)$-dimensional space. In this process, a large set of reference points are sampled on the Pareto front, and then each solution on the resulting Pareto front is replaced by its closest reference point. Thus, a new point set $P'$ can be generated as follows:
\begin{equation}
P' = \{argmin_{r\in R} \parallel - f(x)\parallel | x \in P\},
\end{equation}
where $R$ is a set of reference points and $P$ denotes the set of approximate candidate solutions. After the transformation of $P'$ and $R$ (i.e. projection, translation, and normalization) to project the points to a unit simplex, the ratio of the volume of $P'$ and $R$ is calculated as CPF. 
\begin{equation}
CPF = \frac{Vol(P')} {Vol(R)}.
\end{equation}

The details of this metric and the way of calculating volume are given in \cite{tian2019diversity}.

\textbf{Hausdorff Distance to the Pareto front ($\Delta_p$)} \cite{schutze2012using}: This indicator combines two metrics, GD and IGD. $\Delta_p$ is defined as follows: 

\begin{equation}
\Delta_p(S,P) = max(GD(S,P),IGD(S,P))
\end{equation}
This metric has stronger properties than the two individual indicators since it combines GD and IGD. For instance, $\Delta_p$ can efficiently handle outliers by considering the averaged result. 

\textbf{Pure Diversity (PD)}~\cite{wang2016diversity}: Given an approximate solution $(A)$, PD measures the sum of the dissimilarities of each solution in $A$ to the rest of $A$. For this purpose, the solution with the maximal dissimilarity has the highest priority to accumulate its dissimilarities. The higher the PD metric, the greater the diversity among the solutions. PD is calculated using recursive Eq. \ref{eq10}:
\begin{equation}
\label{eq10}
\begin{aligned}
PD(A)=\max_{s_{i}\in A}(PD(A-s_{i})+d(s_{i},A-s_{i})),
\end{aligned}
\end{equation}
Where,
\begin{equation}
\begin{aligned}
d(s,A)=\min_{s_{i}\in A}(dissimilarity(s,s_{i})),
\end{aligned}
\end{equation}
where $d(s_{i}, X-s_{i})$ denotes the dissimilarity $d$ from one solution $s_{i}$ to a community $A$.

\textbf{Spacing (SP)}~\cite{schott1995fault}: This indicator measures the distribution of non-dominated points on the approximate Pareto front. SP can be computed as follows:
\begin{equation}
SP(S)=\sqrt{\frac{1}{|S|-1}\sum_{i=1}^{|S|}(\overline{d}-d_i)^2},
\end{equation}
where $d_i=\min_{(s_i,s_j)\in S,s_i\neq s_j}\parallel F(s_i)-F(s_j)\parallel_1$ is the $l_1$ distance of $i$th point on the approximate Pareto front to the closest point on the same Pareto front and $\overline{d}$ is the mean of $d_i$'s.

\textbf{Overall Pareto Spread (OS)}~\cite{wang2010multi}: This metric measures the extent of the front covered by the approximate Pareto front. A higher value of this metric indicates better front coverage and it is calculated as follows:
\begin{equation}
OS(S)=\prod_{i=1}^m \frac{|\max_{x \in S}f_i(x)-\min_{x\in S}f_i(x)}{|f_i(P_B)-f_i(P_G)|},
\end{equation}
where $P_B$ is the nadir point and $P_G$ is the ideal point. 

\textbf{Distribution Metric (DM)}~\cite{deb2002running}: This metric also indicates the distribution of the approximate Pareto front by an algorithm. DM is given by:

\begin{equation}
  DM(S)=  \frac{1}{|S|} \sum_{i=1}^{m} \left (\frac{\sigma_i}{\mu_i}\right)\left(\frac{|f_i(P_G)-f_i(P_B)|}{R_i}\right),
\end{equation}
where $\sigma_i$ and $\mu_i$ are the standard deviation and mean of the distances relative to the $i$th objective, $R_i = \max_{ s \in S} f_i(s) - \min_{ s \in S}  f_i(s)$
where $|S|$ is the number of points on the approximate Pareto front, and $f_i(P_G)$ and $f_i(P_B)$ are the function values of design ideal and nadir points, respectively. A lower value of DM indicates well-distributed solutions.

\begin{table*}[htbp]
    \centering
    \caption{Advantages and Disadvantages of Multi-Objective Performance Metrics}
    \label{tab:multi-objective-metrics}
    \begin{tabular}{|p{2.5cm}|p{1.5cm}|p{1cm}|p{5cm}|p{5cm}|}
    \hline
    \multicolumn{1}{|c|}{\textbf{Metric}} & \multicolumn{1}{c|}{\textbf{Measures}} & \multicolumn{1}{c|}{\textbf{Sets}}&  \multicolumn{1}{c|}{\textbf{Advantages}} & \multicolumn{1}{c|}{\textbf{Disadvantages}} \\
    \hline
    Hypervolume (HV) & Accuracy, Diversity & Unary & Provides a single scalar value representing the volume of the objective space covered by the Pareto front. & Computationally expensive for high-dimensional problems. \\
    \hline
    Generational Distance (GD) & Accuracy & Unary & Measures the average distance from solutions in the population to the Pareto front. It tends to be robust to variations in the shape and complexity of the Pareto front. & GD favors solutions that are close to the true Pareto front, potentially overlooking the diversity or spread of solutions in the population. \\
    \hline
    Inverted Generational Distance (IGD) & Accuracy, Diversity & Unary  & Measures the average distance from solutions on the Pareto front to solutions in the population. & Similar to GD, IGD may favor solutions that are close to the true Pareto front, potentially overlooking the convergence or quality of individual solutions. It can also be sensitive to outliers, particularly if extreme solutions significantly affect the average distance calculations. \\
    \hline
    Two-set Coverage (C) & Accuracy, Diversity, Cordiality & Binary & It offers an objective measure of solution quality, enabling comparisons between different algorithms or parameter settings. It emphasizes the importance of diversity by assessing the extent to which the solutions cover the Pareto front. & Requires a reference set for comparison and the quality and characteristics of the reference set can significantly affect the Two-set Coverage metric. Small changes or inaccuracies in the reference set can lead to misleading results. \\
    \hline
    Coverage over the Pareto front (CPF) & Diversity & Unary & Measures the coverage over the Pareto front measures the proportion of the Pareto front covered by a set of solutions. It provides a quantitative assessment of how well the solutions represent the Pareto front. & Sensitive to the density of solutions along the Pareto front. Similar to the Two-set Coverage metric, the quality and characteristics of the reference set can significantly affect the Coverage over the Pareto front metric. Small changes or inaccuracies in the reference set can lead to misleading results. \\
    \hline
    Hausdorff Distance to the Pareto front ($\Delta_p$) & Accuracy, Diversity & Unary & Provides a measure of the maximum distance from any point on the Pareto front to the closest point in a set of solutions. This quantification helps in assessing the quality of the solutions in relation to the Pareto front. & It can be sensitive to outliers, particularly if extreme solutions significantly affect the distance calculations. \\
    \hline
    Pure Diversity (PD) & Diversity & Unary & Quantifies the diversity of solutions in a population without considering their quality or proximity to the Pareto front. This focus on diversity is essential for maintaining a well-distributed set of solutions. & PD does not consider the quality of individual solutions or their proximity to the true Pareto front. Therefore, it may prioritize diversity over convergence or solution quality. \\
    \hline
    Spacing (SP) & Diversity & Unary & provides a measure of the dispersion or spread of solutions in a population. It quantifies how evenly solutions are distributed throughout the objective space. & SP does not directly consider the quality of individual solutions or their proximity to the true Pareto front. Therefore, it may prioritize spread over convergence or solution quality. \\
    \hline
    Overall Pareto Spread (OS) & Diversity & Unary & provides a comprehensive measure of the spread of solutions across the entire Pareto front. It considers the distribution of solutions along both the objective space and the Pareto front. & OS does not directly consider the quality of individual solutions or their proximity to the true Pareto front. Therefore, it may prioritize spread over convergence or solution quality. \\
    \hline
    Distribution Metric (DM) & Diversity & Unary & provides a measure of the distribution or spread of solutions in a population. It quantifies how evenly solutions are distributed throughout the objective space. & DM does not directly consider the quality of individual solutions or their proximity to the true Pareto front. Therefore, it may prioritize spread over convergence or solution quality. It can also be sensitive to the density of solutions in certain regions of the objective space. \\
    \hline
    \end{tabular}
\label{metricsProperties}
\end{table*}


\section{Proposed Pareto-optimal Ranking  Method}
In this section, we present the details of the multi-metric ranking method. The proposed method ranks competing multi- or many-objective algorithms based on a variety of performance metrics, simultaneously. First, each algorithm is independently evaluated using a set of $M$ performance indicators, hence forming $M$ objectives. Then, it combines these M-dimensional points and groups them to Pareto dominance levels using the non-dominated sorting (NDS) algorithm. Finally, the ranks of each algorithm are determined using one of the four proposed ranking methods described in this section. The steps of the proposed method to rank $A$ algorithms (for our experiment we have used $A$ = 10 algorithms) when solving many-objective optimization problems are described below: 

\begin{enumerate}
\item Select $M$-multi-objective performance indicators (e.g., HV, IGD, GD,..., etc.).
\item Run each algorithm $R$ times.
\item For each run, calculate the performance scores based on $M$ metrics. As a result, for each algorithm, we have a matrix size of $R\times M$ performance scores.
\item Concatenate all performance scores from $A$ algorithms to get a matrix size of $A\times R$ points containing $M$-dimensional (objectives/criteria) vectors. 
\item Apply the NDS algorithm on $A\times R$ $M$-dimensional vectors to generate different levels of ranks. Note that, in order to apply the NDS algorithm, we reverse some of the scores (e.g., since the highest HV value indicates the better algorithm, we replace this value with its inverse or change its sign in case of zeros) so that all metrics are evaluated as a minimization problem instead of a maximization.
\item Evaluate the ranks of each algorithm based on their contribution at each Pareto level using the proposed ranking techniques discussed below.
\end{enumerate}
    
Suppose that we want to compare the performance of $A$ multi/many-objective algorithms for solving a benchmark test problem. We run each algorithm $R$ times and compute the algorithm's performance scores using $M$ existing multi-objective metrics (e.g., HV, IGD, GD,..., etc.). After these steps, we obtain $A\times R$ $M$-dimensional points (performance scores).
The format of the matrix representing these scores can be illustrated as follows:
\[
\begin{matrix}
 & \pmb{m_1} & \pmb{m_2} &\pmb{m_3} &\pmb{...}&\pmb{m_M}\\
\pmb{a_{1,1}} & ps_{a_{1,1},m_1} & ps_{a_{1,1},m_2} & ps_{a_{1,1},m_3}&...& ps_{a_{1,1},m_M}\\
\pmb{a_{1,2}} & ps_{a_{1,2},m_1}& ps_{a_{1,2},m_2} & ps_{a_{1,2},m_3}&...& ps_{a_{1,2},m_M}\\
\pmb{\vdots}&\vdots&\vdots&\vdots&\ddots&\vdots&\\
\pmb{a_{1,R}} &ps_{a_{1,R},m_1} & ps_{a_{1,R},m_2} &ps_{a_{1,R},m_3}&...& ps_{a_{1,R},m_M}\\
\pmb{\vdots}&\vdots&\vdots&\vdots&\ddots&\vdots&\\
\pmb{a_{A,1}} &ps_{a_{A,1},m_1} & ps_{a_{A,1},m_2}& ps_{a_{A,1},m_3}&...& ps_{a_{A,1},m_M}\\
\pmb{a_{A,2}} &ps_{a_{A,2},m_1} & ps_{a_{A,2},m_2} & ps_{a_{A,2},m_3}&...& ps_{a_{A,2},m_M}\\
\pmb{\vdots}&\vdots&\vdots&\vdots&\ddots&\vdots&\\
\pmb{a_{A,R}} & ps_{a_{A,R},m_1} & ps_{a_{A,R},m_2} & ps_{a_{A,R},m_3}&...&ps_{a_{A,R},m_M}
\end{matrix}
\]
Where $ps_{a_{i,j}m_k}$ indicates the computed performance score for $j$th run of $i$th algorithm   based on $k$th metric.  

Next, the proposed method applies the NDS algorithm to place $A\times R$ $M$-dimensional vectors. This process yields a set of Pareto levels and the corresponding points in these levels. Suppose that the NDS algorithm resulted in $L$ levels, say $l_1,l_2,l_3,...,l_L$. Then, for each algorithm, we count the number of points associated with each Pareto level (i.e., this step allows us to quantify the quality of each algorithm when solving a given problem). Then, the resulting matrix can be illustrated as follows: 
\[
\begin{matrix}
 & \pmb{l_1} & \pmb{l_2} &\pmb{l_3} &\pmb{...}&\pmb{l_L}\\
\pmb{a_1} & n_{a_1l_1} & n_{a_1l_2} & n_{a_1l_3}&...& n_{a_1l_L}\\
\pmb{a_2} & n_{a_2l_1} & n_{a_2l_2} & n_{a_2l_3}&...& n_{a_2l_L}\\
\pmb{a_3} &n_{a_1l_1} & n_{a_1l_2} & n_{a_1l_3}&...& n_{a_3l_L}\\
\pmb{\vdots}&\vdots&\vdots&\vdots&\ddots&\vdots&\\
\pmb{a_A} & n_{a_Al_1} & n_{a_Al_1} & n_{a_Al_1}&...&n_{a_Al_L}
\end{matrix}
\]
where $n_{a_il_j}$ indicates the number of points (i.e., $M$-dimensional metrics scores) of $i$th algorithm on $j$th Pareto level. Lastly, the ranks of each algorithm are determined using one (or more, in case of a tie) of the four proposed ranking techniques described below. 

\textbf{Olympic method:} The best algorithm is determined by evaluating the number of points each solution has on the first level. If a tie occurs between two solutions, the second level is considered and the algorithm with more points is selected; if there is still a tie, the third level is considered, and so on.
\begin{equation}
   First\_rank= \arg \max_i{(n_{a_il_1})} 
\end{equation}

Suppose that two algorithms, $a_1$ and $a_2$, after the NDS step, have the following number of points on each level:
\[
\begin{matrix}
 & \pmb{l_1} & \pmb{l_2} &\pmb{l_3}\\
\pmb{a_1} & 20 & 10 & 1\\
\pmb{a_2} & 15 &14 & 2
\end{matrix}
\]
where the number of points of algorithm $a_1$ on the first, second, and third levels is 20, 10, and 1, respectively, and 15, 14, and 2 for algorithm $a_2$. According to the Olympic ranking,   algorithm $a_1$ outperforms $a_2$ as the number of points on the first level (i.e., $l_1$) for $a_1$ is higher than $a_2$. 

\textbf{Linear method:} This technique takes all points into account when calculating an algorithm's ranking, rather than just the top Pareto level score like the Olympic method. The weighted score of each algorithm is calculated by multiplying the number of points they have in each level by the decreasing linear weights. For example, if the NDS algorithm produces L levels, the first level will have a weight of $L$, the second level will be $L-1$, and so on. Once the weighted sums are determined for all competing algorithms, their ranks are assigned based on these weights (the highest weighted sum algorithm ranked first). In this way, every point on all levels contributes to the ranking of an algorithm.
Eq.~\ref{LS} represents the computation of linear score for algorithm $a_i$.
\begin{eqnarray}
\begin{aligned}
Linear\_Score(a_i)=&n_{a_il_1}\times (L)+n_{a_il_2}\times (L-1)+,...,\\
&+n_{a_il_L}\times (1)\\
\label {LS}
\end{aligned}
\end{eqnarray}

Given the previous example, the linear score of algorithm $a_1$ and $a_2$ can be calculated as follows:\\
\begin{eqnarray*}
\begin{aligned}
&Linear\_Score(a_1) = 20\times 3 + 10\times 2 + 1\times 1=81\\
&Linear\_Score(a_2) = 15\times 3 + 14\times 2 + 2\times 1=75
\end{aligned}
\end{eqnarray*}
Accordingly, the algorithm $a_1$ has a higher rank.

\textbf{Exponential Method:} Similar to the linear ranking method, the exponential technique assigns a weight to each level of Pareto, however,  the designated weights are decreasing exponentially rather than linearly. Specifically,, the decreasing  weights are $2^{0}$, $2^{-1}$, $2^{-3}$, ..., $2^{-L}$ for levels 1, 2, 3, ..., $L$, respectively. Then, the weighted sum indicates the score of each algorithm. Eq.~\ref{ES} represents the computation of exponential score for algorithm $a_i$.
\begin{eqnarray}
\begin{aligned}
Exponential\_Score(a_i)=&n_{a_il_1}\times 2^{0}+n_{a_il_2}\times 2^{-1}+,...,\\
&+n_{a_il_L}\times 2^{-(L-1)}
\end{aligned}
\label {ES}
\end{eqnarray}
Given the previous example, the exponential score of algorithms $a_1$ and $a_2$ can be calculated as follows:\\
\begin{eqnarray*}
\begin{aligned}
&Exponential\_Score(a_1)  = 20\times2^{0} + 10\times2^{-1} + 1\times2^{-2}\\
&=25.25 \\
&Exponential\_Score(a_2) = 15\times 2^{0} + 14\times 2^{-1} + 2\times 2^{-2}\\
&=22.5
\end{aligned}
\end{eqnarray*}
Accordingly, the algorithm $a_1$ outperforms $a_2$.

\textbf{Adaptive Method:} The score of each algorithm is calculated based on the cumulative number of points distributed over all levels. For each algorithm,  the total number of points at levels 1 and 2 is considered as the cumulative weight of level 2. The total number of points at levels 1, 2, and 3 are considered as the cumulative weight of level 3.  Correspondingly, the total number of points distributed on all levels is the cumulative weight of level $L$ of the corresponding algorithm.  Eq.~\ref{CW} indicates the computation of cumulative weight for level $l$ of algorithm $a_i$.
\begin{eqnarray}
CW(a_i,l)=\sum_{j=1}^{l} n_{a_ij}
\label{CW}
\end{eqnarray} 

Similarly, the total cumulative weight of level $l$ can be defined as:
\begin{eqnarray}
Total\_CW(l)=\sum_{i=1}^{A} CW(a_i,l)
\end{eqnarray}  

Using this definition, the total sum of cumulative weight ratios of all ranks is the score of each algorithm.  In this way, the ratio of contribution of each algorithm at each level determines the rank of the algorithm. Eq.~\ref{AM} represents the computation of adaptive score for algorithm $a_i$.
 \begin{eqnarray}
\begin{aligned}
&Adaptive\_Score (a_i) = \sum_{l=1}^{L} \frac{CW(a_i,l) } {Total\_CW(l) }
\end{aligned}
\label{AM}
\end{eqnarray}  

 For the previous example, we can compute the cumulative weights for algorithms $a_1$ and $a_2$ as follows: 

\[
\begin{matrix}
 & \pmb{CW_1} & \pmb{CW_2} &\pmb{CW_3}\\
\pmb{a_1}& 20 & 30 & 31\\
\pmb{a_2} & 15 &29 & 31
\end{matrix}
\]
The total cumulative weights for each Pareto level are calculated as:
\begin{eqnarray*}
\begin{aligned}
&Total\_CW(l_1)  = 20+15=35 \\
&Total\_CW(l_2)  =30+29=59\\
&Total\_CW(l_3) =31+31=62
\end{aligned}
\end{eqnarray*}  	

Consequently, the adaptive scores for algorithms $a_1$ and $a_2$ will be:
\begin{eqnarray*}
\begin{aligned}
&Adaptive\_Score (a_1)  = 20/35 + 30/59 + 31/62 = 1.58\\
&Adaptive\_Score (a_2) = 15/35+29/59+31/62=1.42
\end{aligned}
\end{eqnarray*}

Thus, based on the adaptive score, algorithm $a_1$ is better than $a_2$. 

 It is worth mentioning that in the event of a tie (i.e., identical ranks for two or more algorithms), when using one of these ranking techniques, we the can apply any of the other ranking methods to break the tie. Additionally, if the user has a preference for ranking in scenarios involving diverse complexities, such as varying numbers of objectives, it is feasible to assign weights to the scores based on their respective complexities following the Non-Dominated Sorting (NDS) step.    

\section{Experimental Validation: Conducting Comprehensive Comparisons}
\subsection{Experimental settings}
The practical application of the proposed metric is utilized to rank ten well-known evolutionary multi-objective algorithms submitted to the 2018 IEEE CEC competition. In this competition, participants were asked to develop a novel many-objective optimization algorithm to solve 15 MaF many-objective test problems listed in Table~\ref{benchmark}. The competing  algorithms include AGE-II~\cite{wagner2013fast}, AMPDEA, BCE-IBEA~\cite{li2015pareto}, CVEA3~\cite{yuan2018cost}, fastCAR~\cite{zhao2018many}, HHcMOEA~\cite{fritsche2018hyper}, KnEA~\cite{zhang2014knee}, RPEA~\cite{liu2017many}, RSEA~\cite{he2017radial}, and RVEA~\cite{cheng2016reference}. All experiments were conducted on 3-,  5-, and 15-objective MAF test problems and the number of decision variables was set according to the setting used in~\cite{cheng2017benchmark}. Each algorithm was run independently 20 times. The maximum number of fitness evaluations was set to $\max (100000, 10000 \times D)$, and the maximum size of the population was set to 240.

In order to assess the efficacy of the proposed ranking method, we obtained the approximated Pareto fronts of these ten algorithms from~\cite{cheng2017benchmark}. The competition utilized the IGD and HV scores to rank these algorithms. However, we have utilized the ten many-objective metrics (including HV and IGD) listed in Section III to take advantage of the different aspects of these performance metrics.

\begin{table}[]
\centering
\caption{Properties of the 15 MaF benchmark problems. $M$ is the number of objectives.}
\setlength{\tabcolsep}{4pt}
\footnotesize
\begin{tabular}{lll}
Test   function & Properties            & Dimension                      \\ \hline
MaF1            & Linear                           &M+9           \\
MaF2            & Concave                            &M+9         \\
MaF3            & Convex, multimodal                 &M+9          \\
MaF4            & Concave, multimodal              &M+9           \\
MaF5            & Convex, biased                    &M+9          \\
MaF6            & Concave, degenerate               &M+9          \\
MaF7            & Mixed, disconnected, multimodal    &M+19         \\
MaF8            & Linear, degenerated                 &2        \\
MaF9            & Linear, degenerated                 &2        \\
MaF11           & Convex, disconnected, nonseparable   & M+9        \\
MaF12           & Concave, nonseparable, biased deceptive   & M+9   \\
MaF13           & Concave, unimodal, nonseparable, degenerate & 5\\ 
MaF14 & Linear, Partially separable,
Large scale
 & 20 $\times$ M \\
MaF15 & Convex, Partially separable,
Large scale & 20 $\times$ M\\\hline
\end{tabular}
\label{benchmark}
\end{table}

\subsection{Ranking algorithms when solving one specific test problem with a particular number of objectives}
Our first experiment includes ranking the ten algorithms based on a specific test problem with a specific number of objectives. Since the number of independent runs is set to 20, the input to our proposed method is 10 $\times$ 20 $M$-dimensional performance metric scores corresponding to each run.    

\begin{figure*}
\centering
\scriptsize
\raisebox{\dimexpr 0.8\baselineskip-\height}
\scriptsize
\begin{tabular}{lccccccc|cccccc}
\hline
\multicolumn{8}{c|}{MaF1 -  M = 5} & \multicolumn{6}{c}{MaF10 -  M = 15} \\ 
\hline
\multicolumn{1}{c}{} & L1 & L2 & L3 & L4 & L5 & L6 & L7 & L1 & L2 & L3 & L4 & L5 & L6 \\ \hline
AGE-II               & 17 & 3  & 0  & 0  & 0  & 0  & 0 & 16 & 2  & 2  & 0  & 0  & 0  \\
AMPDEA               & 4  & 7  & 4  & 4  & 1  & 0  & 0 & 12 & 4  & 2  & 1  & 1  & 0   \\
BCE-IBEA             & 10 & 5  & 4  & 1  & 0  & 0  & 0 & 12 & 3  & 3  & 1  & 1  & 0   \\
CVEA3                & 10 & 5  & 4  & 1  & 0  & 0  & 0 & 20 & 0  & 0  & 0  & 0  & 0    \\
fastCAR              & 15 & 3  & 0  & 0  & 2  & 0  & 0 & 19 & 1  & 0  & 0  & 0  & 0   \\
HHcMOEA              & 19 & 1  & 0  & 0  & 0  & 0  & 0 & 20 & 0  & 0  & 0  & 0  & 0  \\
KnEA                 & 6  & 4  & 5  & 4  & 1  & 0  & 0 & 12 & 3  & 3  & 2  & 0  & 0   \\
RPEA                 & 10 & 5  & 4  & 1  & 0  & 0  & 0 & 12 & 3  & 3  & 1  & 1  & 0   \\
RSEA                 & 9  & 4  & 4  & 2  & 1  & 0  & 0 & 13 & 3  & 3  & 1  & 0  & 0  \\
RVEA                 & 5  & 3  & 4  & 1  & 2  & 4  & 1 & 9  & 4  & 2  & 3  & 1  & 1  \\ \hline
\end{tabular}
\begin{tabular}{c}
{\includegraphics[width=.9\textwidth]{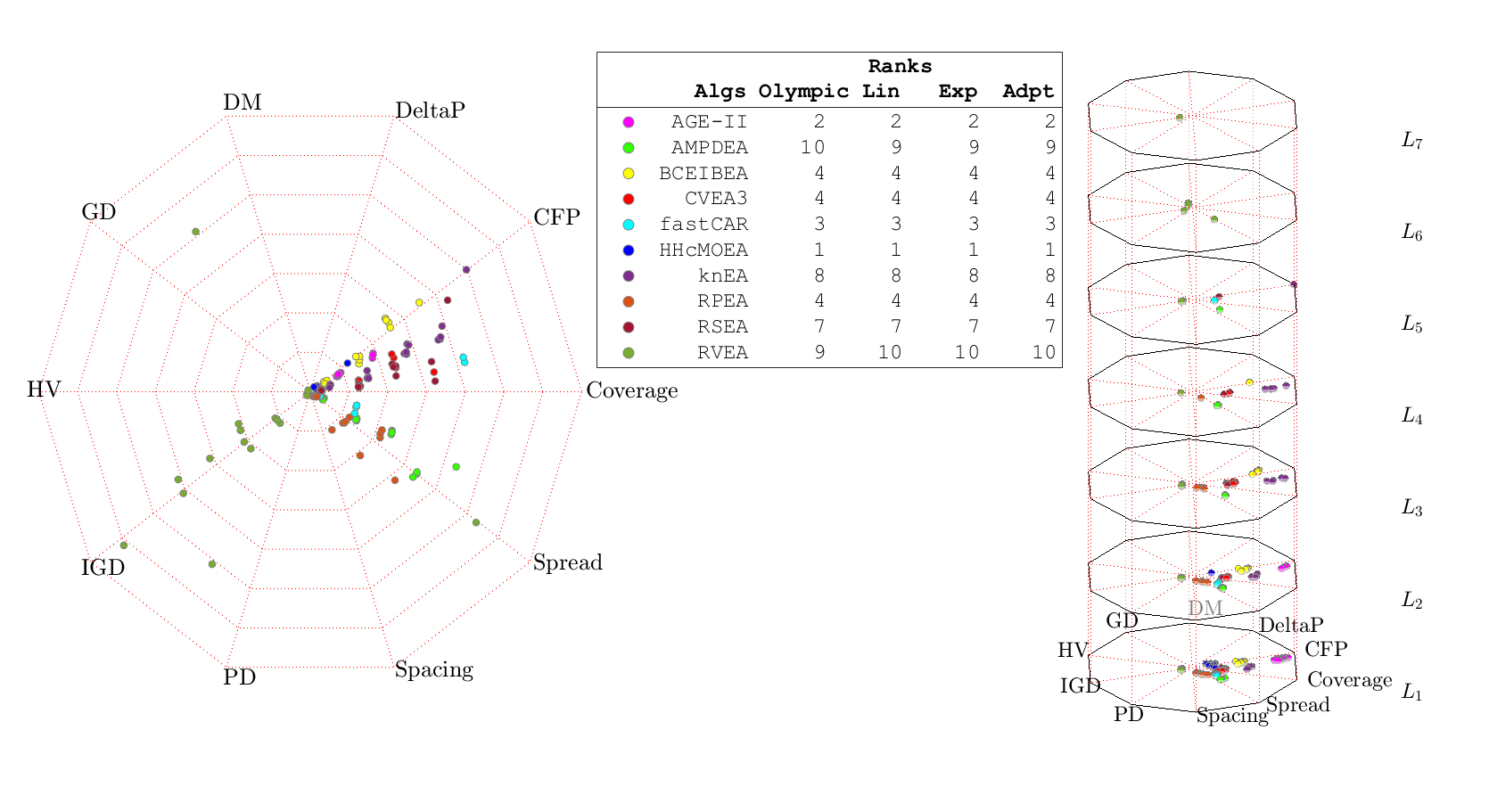}}\\
\small
\textbf{MaF1, M=5}\\
{\includegraphics[width=.9\textwidth]{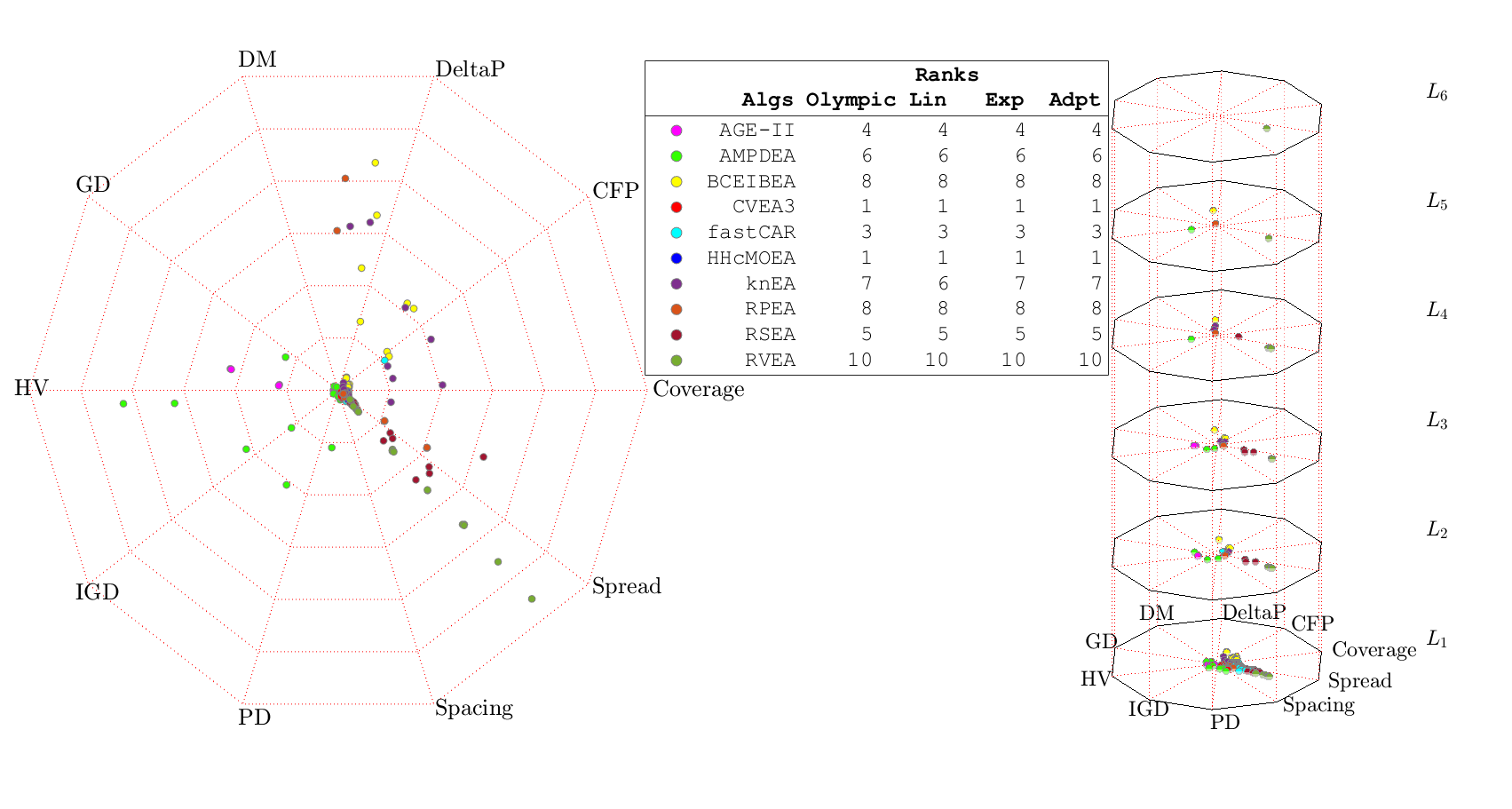}}\\
\small
\textbf{MaF10, M=15}
\end{tabular}

\caption{Outcome of the NDS algorithm and the ranks of algorithms for 5-objective MaF1 and 15-objective MaF10 benchmark test problems. The top table shows the number of points associated with each Pareto level. The bottom two diagrams show the distribution of these points using the PartoRadviz visualization along the ranks of these algorithms using the four ranking techniques proposed in this paper.}
\label{sample}
\end{figure*}

Fig.~\ref{sample} shows the results of the proposed scheme when solving  5-objective MaF1 and 15-objective MaF10 test problems. The NDS algorithm resulted in 7 levels of Pareto levels for MaF1, and 6 Pareto levels for MaF10 test problems. The top table in Fig.~\ref{sample} shows the number of points contributed by each algorithm at each Pareto level. For instance, the scores of  AGE-II for MaF5  resulted in  17 points at the first level, while BCE-IBEA and CVEA3 have 10 points at this level.  Correspondingly, the sum of elements in each row is 20, indicating the overall number of runs.  Fig.~\ref{sample} also illustrates the distribution of these points in different Pareto levels using 2-D and 3-D RadViz visualization~\cite{nasrolahzadeh2020pareto}.  From the figure, the density of points is higher at the lower levels and decreases at the higher Pareto levels. The middle table shows the ranks of each algorithm based on the four ranking methods discussed in section IV.
 For instance, using the Olympic method, the CVEA3 and HHcMOEA algorithms have the highest rank when solving 15-objective MaF10 as all their points are on the first Pareto level. Although the four proposed ranking methods generally provide the same ranking, they may sometimes result in minor conflicts. For example, the AMPDEA algorithm is ranked 10th using the Olympic method, however, it was ranked 9th when the other three ranking methods were used to solve the 5-objective MaF1 test problem. 

Table~\ref{AllFuncs} illustrates the Olympic ranking of each algorithm when addressing the 15-objective MaF1 to MaF15 benchmark test problems. The table reveals instances where the Olympic strategy led to ties among some algorithms. For instance, AMPDEA, CVEA3, and HHcMOEA share the top rank when solving MaF1. In such cases, we can resort to one or more of the other proposed ranking methods or employ the average ranking across all four methods to resolve these ties whenever feasible. However, if two algorithms contribute equally across all Pareto levels, their rankings will remain identical irrespective of the ranking method employed.
\begin{table*}[]
\centering
\caption{Ranks of algorithms when solving 15-objective MaF test problems using the Olympic ranking technique. }
\footnotesize
\setlength{\tabcolsep}{3.3pt}
\begin{tabular}{lcccccccccc}
\hline
\multicolumn{11}{c}{M = 15}                                                                                                                                                                                                                                                                        \\ \hline
      & \multicolumn{1}{l}{AGE-II} & \multicolumn{1}{l}{AMPDEA} & \multicolumn{1}{l}{BCE-IBEA} & \multicolumn{1}{l}{CVEA3} & \multicolumn{1}{l}{fastCAR} & \multicolumn{1}{l}{HHcMOEA} & \multicolumn{1}{l}{KnEA} & \multicolumn{1}{l}{RPEA} & \multicolumn{1}{l}{RSEA} & \multicolumn{1}{l}{RVEA} \\ \hline
MaF1  & 6                          & 1                          & 7                            & 1                         & 10                          & 1                           & 5                        & 4                        & 8                        & 9                        \\
MaF2 & 6                          & 4                          & 8                            & 2                         & 1                           & 7                           & 5                        & 10                       & 3                        & 9                        \\
MaF3 & 6                          & 9                          & 4                            & 7                         & 5                           & 2                           & 1                        & 10                       & 3                        & 8                        \\
MaF4 & 2                          & 5                          & 7                            & 2                         & 2                           & 1                           & 8                        & 6                        & 9                        & 9                        \\
MaF5 & 8                          & 5                          & 3                            & 6                         & 1                           & 10                          & 4                        & 2                        & 9                        & 7                        \\
MaF6 & 5                          & 7                          & 4                            & 7                         & 3                           & 1                           & 1                        & 7                        & 6                        & 7                        \\
MaF7 & 10                         & 4                          & 4                            & 4                         & 3                           & 1                           & 1                        & 4                        & 4                        & 9                        \\
MaF8 & 8                          & 7                          & 1                            & 1                         & 5                           & 9                           & 6                        & 10                       & 3                        & 3                        \\
MaF9 & 10                         & 4                          & 4                            & 4                         & 2                           & 1                           & 4                        & 3                        & 8                        & 9                        \\
MaF10 & 4                          & 6                          & 8                            & 1                         & 3                           & 1                           & 7                        & 8                        & 5                        & 10                       \\
MaF11 & 10                         & 1                          & 9                            & 4                         & 1                           & 3                           & 6                        & 6                        & 8                        & 5                        \\
MaF12 & 6                          & 8                          & 7                            & 2                         & 1                           & 9                           & 2                        & 10                       & 2                        & 2                        \\
MaF13 & 3                          & 5                          & 5                            & 5                         & 1                           & 1                           & 5                        & 5                        & 5                        & 4                        \\
MaF14 & 5                          & 6                          & 4                            & 8                         & 3                           & 1                           & 1                        & 9                        & 10                       & 7                        \\
MaF15 & 1                          & 1                          & 1                            & 1                         & 9                           & 1                           & 7                        & 1                        & 8                        & 10                       \\ \hline
\end{tabular}

\label{AllFuncs}
\end{table*}

\subsection{Ranking algorithms when solving a set of test problems with a particular number of objectives}
\begin{figure*}

\centering
\medskip
\smallskip
\scriptsize
\raisebox{\dimexpr 0.8\baselineskip-\height}
\scriptsize
\setlength{\tabcolsep}{5pt}
\begin{tabular}{lcccccccc|cccccccccccc}
\hline
\multicolumn{9}{c|}{ M = 10} & \multicolumn{12}{c}{  M = 15} \\\hline
\multicolumn{1}{c}{} & L1  & L2 & L3 & L4 & L5 & L6 & L7 & L8& L1  & L2 & L3 & L4 & L5 & L6 & L7 & L8 & L9 & L10 & L11 & L12\\ \hline
AGE-II               & 171 & 63 & 40 & 14 & 10 & 2  & 0  & 0  & 165 & 32 & 17 & 21 & 4  & 32 & 17 & 2  & 0  & 0   & 9   & 1\\
AMPDEA               & 254 & 38 & 8  & 0  & 0  & 0  & 0  & 0 & 190 & 52 & 29 & 16 & 10 & 3  & 0  & 0  & 0  & 0   & 0   & 0 \\
BCE-IBEA             & 272 & 23 & 5  & 0  & 0  & 0  & 0  & 0  & 194 & 46 & 31 & 16 & 10 & 2  & 1  & 0  & 0  & 0   & 0   & 0\\
CVEA3                & 274 & 26 & 0  & 0  & 0  & 0  & 0  & 0 & 207 & 36 & 30 & 14 & 10 & 1  & 1  & 1  & 0  & 0   & 0   & 0  \\
fastCAR              & 250 & 28 & 9  & 11 & 1  & 1  & 0  & 0  & 238 & 43 & 17 & 1  & 0  & 1  & 0  & 0  & 0  & 0   & 0   & 0 \\
HHcMOEA              & 264 & 20 & 13 & 3  & 0  & 0  & 0  & 0  & 232 & 16 & 8  & 4  & 0  & 10 & 12 & 5  & 0  & 0   & 10  & 3 \\
KnEA                 & 246 & 46 & 8  & 0  & 0  & 0  & 0  & 0  & 217 & 37 & 25 & 12 & 7  & 2  & 0  & 0  & 0  & 0   & 0   & 0  \\
RPEA                 & 202 & 61 & 23 & 4  & 6  & 1  & 2  & 1  & 151 & 41 & 28 & 20 & 10 & 18 & 8  & 8  & 7  & 6   & 2   & 1\\
RSEA                 & 233 & 61 & 6  & 0  & 0  & 0  & 0  & 0 & 164 & 40 & 38 & 18 & 12 & 9  & 6  & 4  & 8  & 1   & 0   & 0 \\
RVEA                 & 235 & 22 & 23 & 9  & 11 & 0  & 0  & 0 & 144 & 48 & 45 & 19 & 15 & 10 & 4  & 8  & 4  & 2   & 1   & 0  \\ \hline
\end{tabular}\\
\begin{tabular}{c}
\includegraphics[width=.9\textwidth]{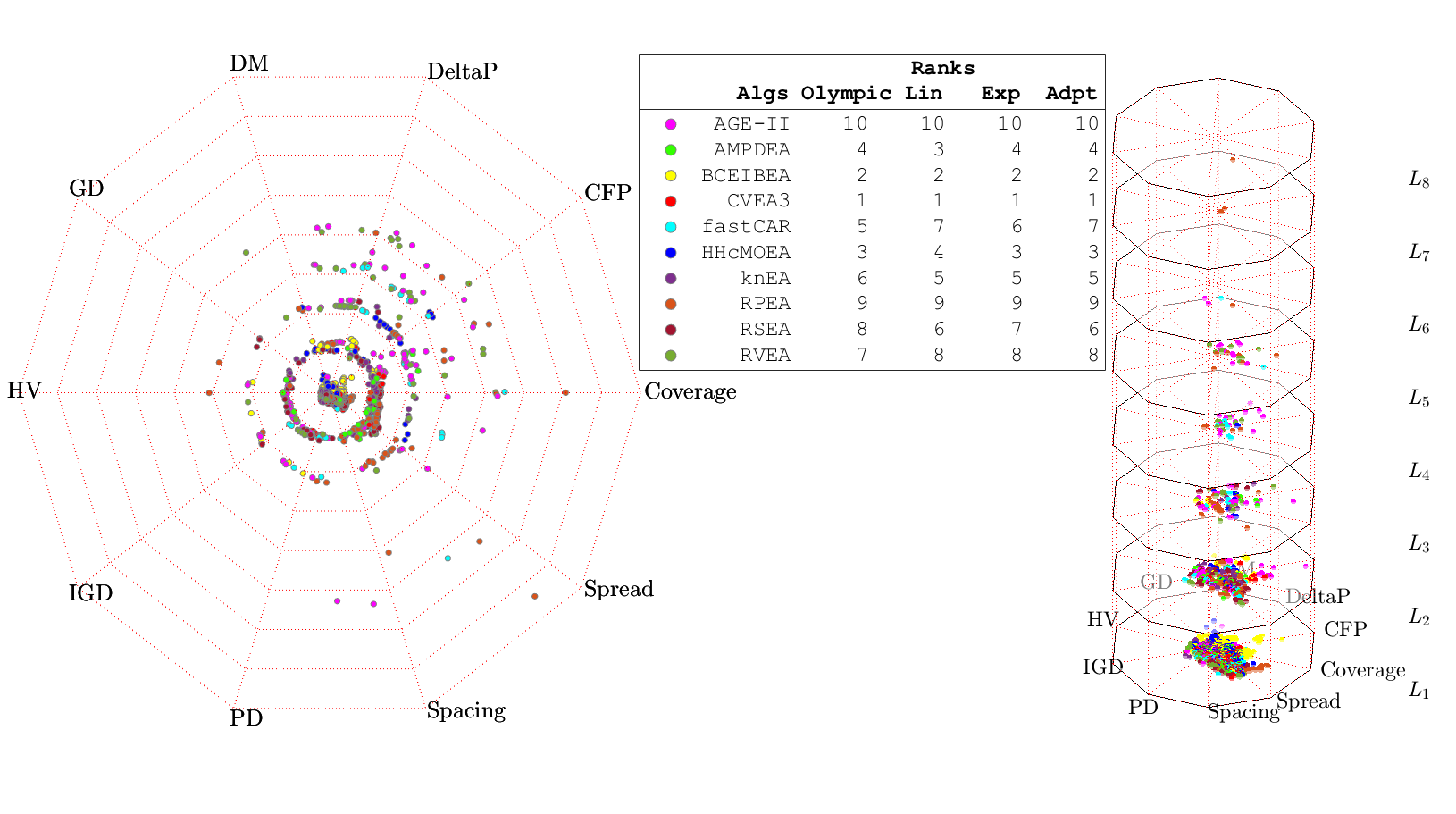}\\
\textbf{M=10}\\
\includegraphics[width=.9\textwidth]{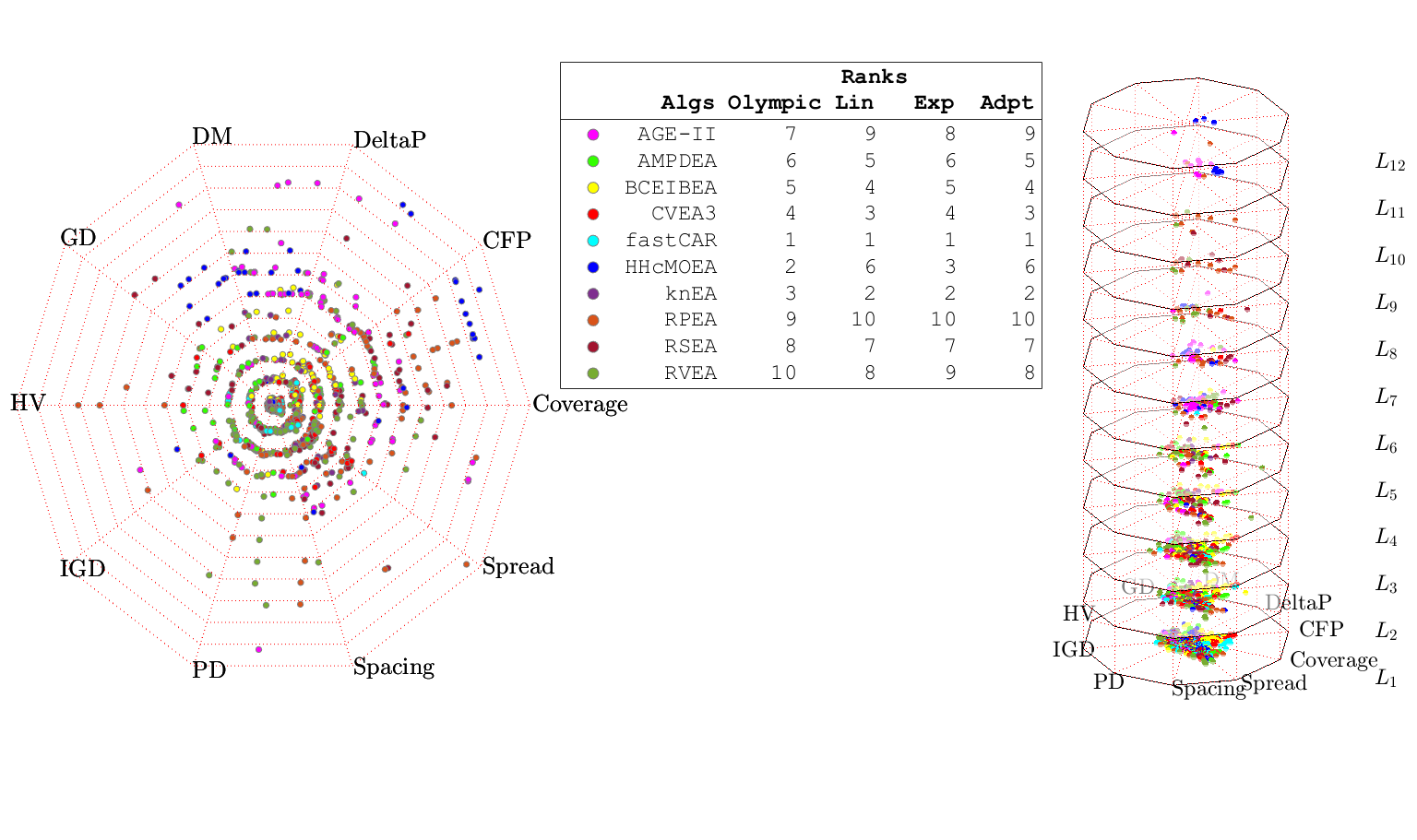}\\
\textbf{M=15}\\
\end{tabular}
\medskip
\caption{The overall outcome of the NDS algorithm and the ranks of the ten algorithms when solving 10- and 15-objective MaF benchmark test problems. The top table shows the total number of points associated with each Pareto level. The bottom two diagrams show the distribution of these points using the PartoRadviz visualization along the ranks of these algorithms using the four ranking techniques proposed in this paper.}
\label{Allfunctions}
\end{figure*}

In the previous experiment, we examined the ranking of algorithms based on individual test problems with a specific number of objectives. In this experiment, we delve into the overall ranking of algorithms across a set of test problems with the same number of objectives. To establish the overall rankings of algorithms, akin to the previous experiment, we initially compute the contribution of each algorithm to each specific test problem. Subsequently, for each algorithm, we aggregate the total number of points at each level across all test problems with the same index. For instance, if Algorithm 1 accumulates 20 points in level 1 when addressing test problem 1, and 12 and 8 points in levels 1 and 2, respectively, when addressing test problem 2, then the total number of points for Algorithm 1 would be 32 in level 1 and 8 in level 2. The table in Fig.  \ref{Allfunctions}  represents each algorithm's total number of points on each Pareto level when solving 10- and 15-objective MaF benchmark test problems. For instance, AGE-II contributed 171 points on the first level from all test problems when solving 10-objective MaF test problems. From the table, we can also observe that the sum of points in each row is 300 (i.e., 300 = 20 runs $\times$ 15 test problems).  

 Fig. \ref{Allfunctions}  also shows the overall distribution of points on the different Pareto levels using 2-D and 3-D RadViz visualization. Additionally, the rankings of each algorithm based on all test problems using the four proposed ranking schemes can be seen in the middle table of Fig.~\ref{Allfunctions}. For instance, when solving 15-objective MaF test problems, the fastCAR algorithm has the maximum number of points on the first level compared to the other algorithms. As a result, it is ranked first according to the Olympic method as well as the other ranking methods.  On the other hand, the HHcMOEA algorithm is ranked second by the Olympic method when solving 15 objective test problems as it has the second-highest number of NDS contributions at the first level.  However, this algorithm is ranked 6th when using the linear method to rank these algorithms. Hence, the ranking methods should be selected carefully to address the needs of each user. For example, the Olympic method would be useful when we are only interested in algorithms that have the majority of their contribution in Pareto level 1.   


\subsection{Determining the overall rankings of algorithms}
In order to determine the overall rankings of algorithms, it is required to consider the results of all algorithms when solving all test problems with a different number of objectives. From the previous experiment, we have identified the contribution of each algorithm after the NDS step when solving 5-, 10-, and 15-objective MaF test problems separately. Now, we combine these contributions by adding the number of points at each level with the same index from the results of the three different objectives (when $M$ = 5, 10, and 15). Hence, we have a total of nine hundred 10-dimensional points (15 test problems $\times$ 3 different numbers of objectives $\times$ 20 runs). For example, suppose we are interested in ranking algorithms based on their performance when $M$ = 5 and 10.  Let Algorithm 1 has 120 points in level 1 and 80 points in level 2 when solving test problems with $M$=5, and 100 points in level 1, 55 points in level 2, and 45 in level 3 when solving test problems with $M$=10. Then, the total number of points of Algorithm 1 would be 220 in level 1, 135 in level 2, and 45 in level 2.        
 
 From Table~\ref{Overall} we see that the combined results consist of 18 ranks, and each algorithm consists of a total of 900 points. For some of the algorithms, such as AMPDEA, most of the points are located on the higher levels (i.e., the first 5 levels) while for some others such as RVEA, the points are distributed over 18 levels. Table~\ref{Overallrank} shows the overall ranking of all algorithms using the four different ranking techniques discussed in Section IV along with their average ranking based on the four ranking techniques. For instance, the AGE-II algorithm is ranked 5th position using the Olympic method while this algorithm takes the 7th position if the linear ranking method is applied. However, for some algorithms such as RSEA an equal rank is assigned by all techniques. Based on the results obtained from the proposed four ranking methods, the fastCAR algorithm is ranked 1st algorithm among the 10 state-of-the-art many-objective optimization algorithms based on the ten considered performance indicators, while RPEA, conversely is ranked last.

It is important to highlight that all four proposed ranking methods produce satisfactory results. However, we suggest employing the average ranking across these methods to impartially evaluate them based on a comprehensive assessment. Additionally, using the average ranking helps mitigate discrepancies that may arise when different methods assign varying rankings to the same algorithm. Alternatively, if one prefers to use a single ranking method, we recommend utilizing the adaptive ranking method for the following reasons. Firstly, unlike the linear ranking method, it considers the cumulative points distributed across all levels, as opposed to only considering the top-level points. Secondly, although all the proposed ranking methods demonstrated comparable ranking performance, the adaptive ranking method exhibited a higher average pair-wise correlation value compared to the other methods. This indicates that the results obtained from the adaptive ranking method are more consistent with the rankings generated by other methods. The average pair-wise correlation values for the Olympic, linear, exponential, and adaptive ranking methods are 0.947, 0.923, 0.956, and 0.960, respectively. Another consideration is that the Olympic method is particularly useful for identifying algorithms whose primary contributions lie within the top Pareto level/s.

\begin{table*}[]
\caption{Distribution of 900 points from each algorithm over 18 Pareto levels. }
\setlength{\tabcolsep}{4.1pt}
\footnotesize
\centering
\begin{tabular}{lcccccccccccccccccc}
\hline
\multicolumn{1}{c}{} & L1  & L2  & L3  & L4 & L5 & L6 & L7 & L8 & L9 & L10 & L11 & L12 & L13 & L14 & L15 & L16 & L17 & L18 \\ \hline
AGE-II               & 601 & 127 & 59  & 35 & 15 & 34 & 17 & 2  & 0  & 0   & 9   & 1   & 0   & 0   & 0   & 0   & 0   & 0   \\
AMPDEA               & 571 & 166 & 97  & 40 & 19 & 6  & 1  & 0  & 0  & 0   & 0   & 0   & 0   & 0   & 0   & 0   & 0   & 0   \\
BCE-IBEA             & 618 & 134 & 89  & 37 & 14 & 5  & 2  & 1  & 0  & 0   & 0   & 0   & 0   & 0   & 0   & 0   & 0   & 0   \\
CVEA3                & 619 & 136 & 86  & 33 & 13 & 4  & 4  & 3  & 2  & 0   & 0   & 0   & 0   & 0   & 0   & 0   & 0   & 0   \\
fastCAR              & 747 & 83  & 31  & 15 & 4  & 3  & 0  & 3  & 0  & 3   & 7   & 1   & 0   & 0   & 3   & 0   & 0   & 0   \\
HHcMOEA              & 747 & 53  & 30  & 12 & 8  & 15 & 13 & 8  & 1  & 0   & 10  & 3   & 0   & 0   & 0   & 0   & 0   & 0   \\
KnEA                 & 595 & 142 & 86  & 37 & 13 & 4  & 4  & 4  & 4  & 4   & 3   & 2   & 1   & 1   & 0   & 0   & 0   & 0   \\
RPEA                 & 467 & 152 & 102 & 60 & 43 & 29 & 13 & 13 & 9  & 8   & 3   & 1   & 0   & 0   & 0   & 0   & 0   & 0   \\
RSEA                 & 517 & 165 & 99  & 48 & 28 & 17 & 11 & 6  & 8  & 1   & 0   & 0   & 0   & 0   & 0   & 0   & 0   & 0   \\
RVEA                 & 513 & 135 & 125 & 44 & 29 & 14 & 5  & 8  & 4  & 2   & 1   & 5   & 2   & 2   & 4   & 4   & 2   & 1   \\ \hline
\end{tabular}
\label{Overall}
\end{table*}

\begin{table}[]
\caption{The overall ranking of algorithms using the four ranking techniques based on their contribution presented in Table~\ref{Overall}  }
\footnotesize
\centering
\begin{tabular}{lccccc}
\hline
\multicolumn{1}{c}{} & Olympic & Linear & Expo & Adaptive & Average Rank\\ \hline
AGE-II               & 5       & 7      & 6    & 7   & 7        \\
AMPDEA               & 7       & 5      & 7    & 5   & 6        \\
BCE-IBEA             & 4       & 2      & 4    & 3   & 3        \\
CVEA3                & 3       & 4      & 3    & 4   & 4        \\
fastCAR              & 1       & 1      & 1    & 1   & 1        \\
HHcMOEA              & 2       & 3      & 2    & 2   & 2        \\
KnEA                 & 6       & 6      & 5    & 6   & 5        \\
RPEA                 & 10      & 10     & 10   & 10   & 10       \\
RSEA                 & 8       & 8      & 8    & 8   & 8       \\
RVEA                 & 9       & 9      & 9    & 9   & 9        \\ \hline
\end{tabular}

\label{Overallrank}
\end{table}

\subsection{Comparison of rankings from the Competition and the proposed method}
In this section, we evaluate the results from the proposed ranking method and the official rankings published by the CEC 2018 competition when comparing ten evolutionary multi- and many-objective optimization algorithms on 15 MaF benchmark problems with 5, 10, and 15 objectives. Each algorithm was run independently 20 times for each test problem with 5, 10, and 15 objectives to generate 900 results (15 test problems $\times$ 3 different numbers of objectives $\times$ 20 runs). The Committee ranked the 10 algorithms based on two multi-objective metrics, IGD and HV. They sorted the means of each performance indicator value on each problem with each number of objectives (i.e., 90 ranks). The score achieved by each algorithm was then computed as the sum of the reciprocal values of the ranks. 

For a fair comparison with the official ranking provided by the ECE2018 committee, in this experiment, we have used the HV and IGD metrics (as opposed to the ten metrics used in the previous experiments) to rank these algorithms using the proposed ranking methods. Table~\ref{Comp} represents the results of ranking based on the competition scores and the proposed method. From this table, we see that, since both ranking methods use the HV and IGD metrics, both methods resulted in comparable rankings, with CVEA3 as 1st and RPEA as last. However, when comparing the results obtained by the proposed algorithm when utilizing the ten performance metrics and the official CEC2018 results provided by the committee, we see a significant difference in the rankings of these algorithms. This is expected as the proposed ranking method uses several metrics to evaluate the rankings of each algorithm based on their performance in all aspects (convergence, distribution, spread, ..., etc.) of many-objective quality measures. For instance, the proposed method ranked the fastCAR algorithm 1st when evaluating the overall performance of algorithms based on ten performance metrics while RPEA is ranked last. 
From the above experiments, we see the importance of incorporating several performance metrics to properly assess the performance of multi-/many-objective algorithms in all aspects of their quality measures for the following reasons:
\begin{itemize}
\item
A single metric cannot capture all aspects of algorithm performance: Different metrics measure different aspects of algorithm performance, and no single metric can capture all of them. Using multiple metrics gives a more comprehensive picture of how the algorithms perform in different areas.
\item
We need at least as many indicators as the number of objectives in order to determine whether an approximate solution is \emph{better} than another~\cite{zitzler2003performance}. 
\item
Metrics may contradict each other: different metrics may indicate different levels of performance for the same algorithm, and, in some cases, metrics may even contradict each other. By comparing algorithms with multiple metrics, you can get a more nuanced understanding of their strengths and weaknesses.
\item
It can help avoid bias: using a single metric to compare algorithms can introduce a bias, especially if the metric is not representative of the problem being solved. By using multiple metrics, you can reduce the risk of bias and get a fair objective evaluation of algorithms.
\item
The proposed scheme is flexible: the proposed schemes allow researchers to include/exclude existing or future metrics as part of their contributing ranking metrics. 
\item
The proposed scheme is scalable: it is possible to rank algorithms for an increased number of algorithms, metrics, and number of runs without any change or introducing additional parameters. 

\end{itemize}

Overall, using multiple metrics to compare algorithms provides a more comprehensive and accurate evaluation of their performance, and can help us to make more informed decisions when choosing between them. Therefore, the proposed ranking method can help us incorporate several performance metrics to adequately rank multi-/ many-objective algorithms based on their overall achievement in several categories of performance measures. While our experiments haven't revealed any significant problem with the proposed ranking schemes, it's essential to recognize that the suggested ranking method could be exploited if someone customizes a competing algorithm to excel in a specific indicator, thus becoming the top algorithm according to the Pareto dominance definition used in this paper. In order to mitigate these concerns, we recommend employing impartial domination relations, such as $\varepsilon$-dominance ~\cite{kang2007new}, to ensure that none of the competing algorithms exploit the limitation inherent in the NDS algorithm. The $\varepsilon$-dominance is defined as follows:

Given two vectors, $\pmb x=(x_{1},x_{2},...,x_{d})$ and  $ \acute{\pmb x}=(\acute{x}_{1},\acute{x}_{2},...,\acute{ x}_{d})$ in a minimization problem search space, we say that $\pmb x$ $\varepsilon$-dominates $\acute{\pmb x}$ ($\pmb x\prec_{\varepsilon}\acute{\pmb x}$) if and only if $(B_t - W_s = \varepsilon > 0) \wedge (||F(\pmb x)|| < ||F(\acute{\pmb x)}||)$, where

$B_t(\pmb x, \acute{\pmb x})$ = $|{F_i(\pmb x) < F_i(\acute{\pmb x})}|$, $i\in M$

$W_s(\pmb x, \acute{\pmb x})$ = $|{F_i(\pmb x) > F_i(\acute{\pmb x})}|$, $i\in M$

$||F(\pmb x)|| = \sqrt{\sum\limits_{i=0}^{m} (F_i(\pmb x))^2}$

and when $W_s =0$, $\varepsilon$-dominance is equivalent to Pareto dominance. 

\begin{table*}[]
\caption{Comparison between the ranking results provided by the Competition and the proposed techniques based on HV and IGD measures.}
\footnotesize
\centering
\begin{tabular}{lcccccc}
\hline
\multirow{2}{*}{Algorithm} & \multicolumn{5}{c}{Proposed method} & \multirow{2}{*}{Competition ranking} \\ 
\cline{2-6}
\multicolumn{1}{c}{}                           & Olympic  & Linear & Expo & Adaptive &  Average Rank &                                     \\ \hline
AGE-II                                         & 4        & 5      & 4    & 6    & 4        & 5                                \\
AMPDEA                                         & 2        & 2      & 2    & 2    & 2        & 2                                    \\
BCE-IBEA                                       & 3        & 4      & 3    & 3    & 3        & 3                                    \\
CVEA3                                          & 1        & 1      & 1    & 1    & 1        & 1                                    \\
fastCAR                                        & 7        & 3      & 7    & 4    & 5       & 6                                    \\
HHcMOEA                                        & 10       & 7      & 10   & 8    & 9        & 8                                    \\
KnEA                                           & 6        & 9      & 6    & 7    & 7        & 7                                    \\
RPEA                                           & 9        & 10     & 8    & 10    & 10       & 10                                   \\
RSEA                                           & 5        & 6      & 5    & 5    & 5        & 4                                    \\
RVEA                                           & 8        & 8      & 9    & 9    & 8        & 9                                    \\ \hline
\end{tabular}
\label{Comp}
\end{table*}

\section{Conclusion}
In this study, we have proposed a novel ranking technique to assess the quality of many-objective optimization algorithms using a set of performance metrics. Since there is no one many-objective performance indicator that can capture all quality aspects of an algorithm (convergence, distribution, spread, cardinality, ..., etc.), it is important to incorporate several performance multi-/many-metrics to properly assess the performance of multi-/many-objectives algorithms in all aspects of many-objective quality measures. The proposed multi-metric approach gives users the ability to incorporate as many performance indicators as required to properly compare the quality of several competing algorithms. The proposed ranking method uses the NDS algorithm to categorize the level of contribution from each algorithm and apply four ranking techniques, namely, Olympic, linear, exponential, and adaptive to rank algorithms based on several performance metrics. Our experimental results indicate that the proposed ranking method can effortlessly incorporate several performance metrics to adequately rank multi-/many-objective algorithms based on their overall achievements in several categories of performance measures. Moreover, it can also be used as a general ranking technique for any application in which the evaluation of multiple metrics is required. This includes such as machine learning (e.g., multi-loss), data mining (e.g., multi-quality metrics), business (e.g., revenue, profitability, customer satisfaction, employee engagement, market share), sport (e.g., scoring, assists, rebounds, blocks, tackles), healthcare (e.g., blood pressure, cholesterol levels, body mass index, heart rate variability, cognitive function), education (e.g., grades, standardized test scores, attendance), and environment (e.g., air quality, water quality, biodiversity, and climate change).

\bibliographystyle{IEEEexample}
\bibliography{IEEEexample.bib}

\end{document}